\pdfoutput=1

\documentclass[11pt]{article}

\usepackage[final]{acl}

\usepackage{microtype}
\usepackage{graphicx}
\usepackage{subfig} 
\usepackage{float} 
\usepackage{subfig}
\usepackage{booktabs} 
\usepackage[most]{tcolorbox} 
\usepackage{arydshln}
\usepackage{hyperref}
\usepackage{multirow}
\usepackage{colortbl} 
\usepackage{graphicx}
\usepackage{multirow}
\usepackage{subfig}
\usepackage{amsmath}
\usepackage[most]{tcolorbox}
\definecolor{darkgreen}{RGB}{0, 100, 0}
\usepackage{enumitem}
\usepackage[utf8]{inputenc} 
 
\definecolor{LightCyan}{rgb}{0.88,1,1}

\newcounter{myboxcounter}
\newtcolorbox[auto counter, number within=none]{mybox}[1][]{
  enhanced,
  breakable,
  fonttitle=\bfseries,
  title={Example: Caption and Hard Negative Caption \themyboxcounter},  
  colframe=green!50!black,  
  colback=green!5!white,    
  coltitle=black,           
  label={ex:\themyboxcounter}, 
  #1
}

\usepackage{amsmath}
\usepackage{amssymb}
\usepackage{mathtools}
\usepackage{amsthm}
\usepackage{makecell} 
\usepackage[capitalize,noabbrev]{cleveref}
\usepackage[normalem]{ulem}
\usepackage{svg}

\usepackage{times}
\usepackage{latexsym}

\usepackage[T1]{fontenc}

\usepackage[utf8]{inputenc}

\usepackage{microtype}

\usepackage{inconsolata}

\usepackage{graphicx}

\title{On the Perception Bottleneck of VLMs for Chart Understanding}

\author{Junteng Liu$^{1}$, Weihao Zeng$^{1}$, Xiwen Zhang$^{2}$, Yijun Wang$^{3}$, Zifei Shan$^{3},$ Junxian He$^{1}$\\
  $^{1}$The Hong Kong University of Science and Technology,\\
  $^{2}$Independent contributor,
  $^{3}$Tencent \\
  \texttt{jliugi@cse.ust.hk},
  \texttt{junxianh@cse.ust.hk} \\}

\begin{document}
\maketitle
\begin{abstract}
Chart understanding requires models to effectively analyze and reason about numerical data, textual elements, and complex visual components. Our observations reveal that the perception capabilities of existing large vision-language models (LVLMs) constitute a critical bottleneck in this process. In this study, we delve into this perception bottleneck by decomposing it into two components: the vision encoder bottleneck, where the visual representation may fail to encapsulate the correct information, and the extraction bottleneck, where the language model struggles to extract the necessary information from the provided visual representations.
Through comprehensive experiments, we find that (1) the information embedded within visual representations is substantially richer than what is typically captured by linear extractors, such as the widely used retrieval accuracy metric;
(2) While instruction tuning effectively enhances the extraction capability of LVLMs, the vision encoder remains a critical bottleneck, demanding focused attention and improvement.
Therefore, we further enhance the visual encoder to mitigate the vision encoder bottleneck under a contrastive learning framework.  
Empirical results demonstrate that our approach significantly mitigates the perception bottleneck and improves the ability of LVLMs to comprehend charts. Code is publicly available at \url{https://github.com/hkust-nlp/Vision4Chart}
\end{abstract}

\section{Introduction}

\begin{figure*} [t]
    \centering
    \includegraphics[width=0.85\textwidth]
    {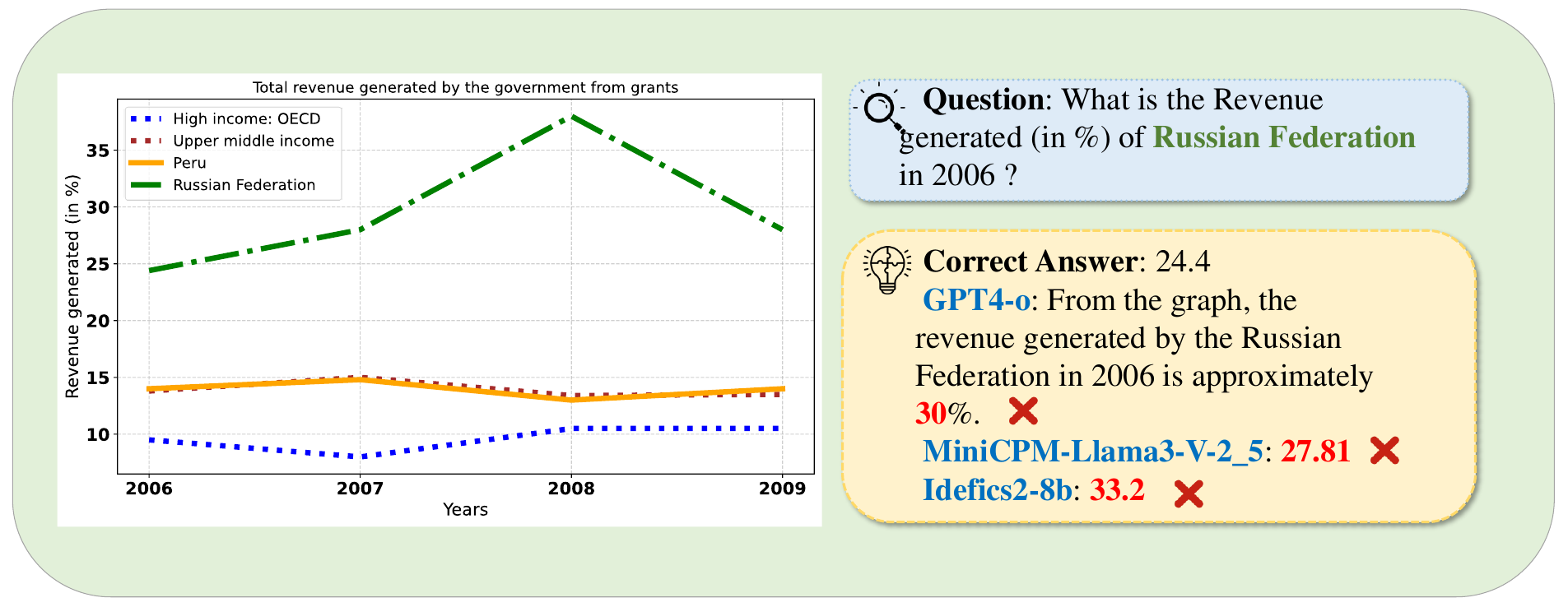}

    \caption{An example of perception QA from the PlotQA dataset~\citep{methani2020plotqa}, along with the responses from GPT4-o~\citep{achiam2023gpt}, MiniCPM~\citep{yao2024minicpmv}, and Idefics2~\citep{laurenccon2024matters} for this example. (The chart has been redrawn for clarity in presentation.)}

    \vspace{-15pt}
    \label{fig-example}
\end{figure*}

Charts are essential for representing and analyzing data, commonly appearing in scientific papers, financial reports, and news articles. Unlike natural images, where semantic content is often apparent through object recognition, charts encode dense quantitative and relational information via visual elements such as bars, lines, and points, along with their spatial relationships. This information-dense property creates higher perception challenges for large vision-language models (LVLMs), which, despite success in general visual understanding tasks \citep{alayrac2022flamingo, li2023blip, liu2024visual}, often struggle with chart understanding \citep{masry-etal-2022-chartqa, ChartBench, xia2024chartx, wang2024charxiv, huang-etal-2024-chart}, as demonstrated in Figure~\ref{fig-example}.  

In this work, we systematically examine the perception bottleneck of LVLMs by analyzing it through two key components: the vision encoder and the language model. Specifically, we define perception as the model's ability to accurately extract visual information from an image. For most LVLMs equipped with a dedicated vision encoder, the process of perceiving visual signals can be broken down into two stages: first, the vision encoder encodes the image into compact vector representations; second, the language model extracts the relevant information from these encoded vectors.
Accordingly, we decompose the perception bottleneck into two categories: the \textbf{vision encoder bottleneck} and the \textbf{extraction bottleneck}. Our objective is to investigate how these two distinct bottlenecks impact the overall perception capability of LVLMs and how to mitigate them.

We begin our study of the vision encoder bottleneck by evaluating the chart understanding ability of CLIP, a widely used vision encoder in many LVLMs~\citep{liu2024improved,liu2024visual,laurenccon2024matters}. Specifically, we construct an image-text retrieval test set using existing chart-specific datasets and assess CLIP with the standard retrieval accuracy.
We find the CLIP performs nearly random retrieval accuracy on these chart datasets, which suggests it may experience significant information loss, as several prior studies use CLIP's retrieval accuracy as an indicator of the information contained in its visual embeddings~\citep{tong2024eyes,deng2024seeing}.
While some researchers attribute this failure to CLIP's inductive bias or intrinsic limitations \citep{tong2024eyes,kamath2023whats}, we successfully develop enhanced CLIP models with substantially improved retrieval accuracy. Specifically, we fine-tune CLIP on chart-specific datasets within a contrastive learning framework and incorporate hard negative captions~\citep{negclip2022}. 
The gains of over 20 absolute points in our enhanced CLIP strongly suggest that CLIP can indeed learn subtle or non-semantic features through further contrastive learning.

To investigate the extraction bottleneck in the language model part, we shift our focus to LVLMs built on top of these CLIP vision encoders. Specifically, we conduct LLaVA-style training~\citep{liu2024improved} combined with chart-specific instruction tuning. Our initial observations reveal that LVLMs trained with the LLaVA data perform poorly on chart understanding tasks, while achieving substantial improvement further fine-tuned on chart-specific data, even with the vision encoder kept frozen.
This finding not only indicates that domain-specific instruction tuning effectively addresses the extraction bottleneck, but more interestingly, it suggests that poor CLIP retrieval accuracy does not necessarily indicate a lack of useful encoded information. 

In contrast, evaluating across seven chart-related benchmarks, spanning both in-distribution and out-of-distribution scenarios, our enhanced CLIPs-based LVLMs further achieve larger gains due to the mitigation of the vision encoder bottleneck. Notably, compared to the original CLIP-based LVLMs, the enhanced CLIP-based models using the LLaVA-v1.5-13B architecture achieve an average improvement of nearly 3 points, while the model employing the LLaVA-v1.5-Phi-3.8B architecture demonstrates an even more significant improvement of 5 points.

Finally, we conduct an in-depth analysis to understand how the superior performance of CLIP translates to its LVLM counterpart. By scaling instruction tuning on larger chart datasets and analyzing CLIP-LLaVA correctness statistics, we observe that samples correctly classified by CLIP are more easily learned by the LVLM, suggesting that the more salient representations obtained from the enhanced CLIP facilitate better LVLM learning. These findings further raise rethinking about information encoding in CLIP and its effect on LVLMs.

\begin{figure*} [t]
    \centering
    \includegraphics[width=0.9\textwidth]
    {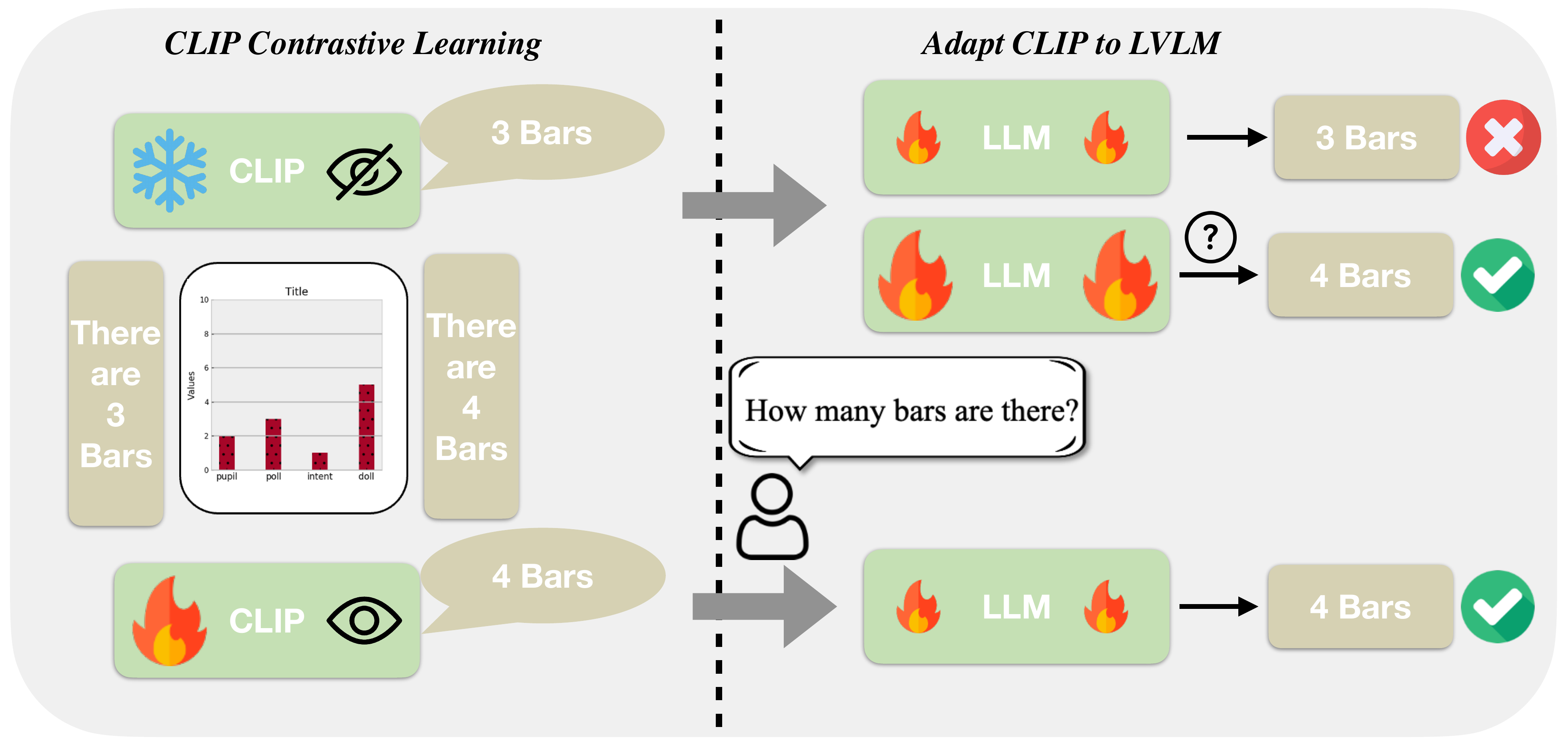}

\caption{\textbf{Left}: A CLIP-blind case where the original CLIP fails to discriminate the number of bars in the chart. By leveraging contrastive learning with hard negatives, the enhanced CLIP model learns more discriminative visual features successfully. \textbf{Right}: When adapted to LVLMs, after instruction tuning, the original CLIP-LVLMs are possible to correctly interpret the chart information even when the original CLIP fails to discriminate it. However, the enhanced CLIP-LVLMs enable faster learning and achieve higher overall performance.}
    \label{fig-main}
\end{figure*}

\section{The Challenge of Chart Understanding}

To better illustrate the perceptual challenges in chart understanding, we examine one concrete perception QA example from PlotQA~\citep{methani2020plotqa}. As shown in Figure~\ref{fig-example}, to answer ``What is the Revenue generated (in \%) of Russian Federation in 2006?'', models need to: (1) correctly match the dotted green line with its legend label, (2) locate the intersection point between this line and the vertical line at 2006, and (3) accurately map this point to the y-axis scale to obtain the value ($\sim$  24\%). While humans can perform this visual reasoning process effortlessly, current models like GPT4-o~\citep{achiam2023gpt}, MiniCPM~\citep{yao2024minicpmv} and Idefics2~\citep{laurenccon2024matters} often struggle with such perception tasks as demonstrated in Figure~\ref{fig-example}.
Unlike natural images, chart understanding presents unique perception challenges as it requires accurately encoding and processing dense quantitative information encoded in visual elements.

Recent studies have quantitatively revealed these perceptual limitations through new chart-specific benchmarks~\citep{ChartBench,wang2024charxiv,xia2024chartx}.  To better understand the sources of these limitations, we decompose the perception bottleneck into two key components.
(1) The vision encoder bottleneck: This occurs when the vision encoder fails to encode critical information from the image into its embeddings, leading to inevitable failures in downstream LVLM tasks. (2) The extraction bottleneck: Even when the image embeddings contain the necessary information, the LLM struggles to extract and interpret them correctly, resulting in erroneous outputs.
In our study, we investigate the impact of these two bottlenecks and propose strategies to mitigate them on the chart understanding task.
Next, we start by analyzing the vision encoder bottleneck.

\section{The Vision Encoder Bottleneck: Investigating and Improving CLIP}

As CLIP~\citep{radford2021learning} serves as the vision encoder in most LVLMs~\citep{liu2024improved,liu2024visual,laurenccon2024matters,zhu2024llava}, we focus on CLIP to investigate the vision encoder bottleneck. In this section, we construct a framework for training and evaluating CLIP's chart understanding abilities. 

\subsection{Background of CLIP}
The CLIP model consists of an image encoder and a text encoder, which map paired image and text data into corresponding vector representations. It employs contrastive learning to align these representations in a shared embedding space. The training objective maximizes the similarity between matched image-text pairs while minimizing it for unmatched pairs, effectively bridging visual and textual modalities for robust cross-modal understanding.

\begin{table*}[t]
\caption{Image-to-Text retrieval evaluation accuracy on original CLIP-ViT-L/14-336px and fine-tuned CLIPs. DVQA-E indicates DVQA Easy, and DVQA-H indicates DVQA Hard. Improvements in the ``Avg.'' column are marked with \textcolor[HTML]{228B22}{$\uparrow$} compared to the CLIP baseline.}
\label{tab:original-clip-eval}
\begin{center}
\renewcommand{\arraystretch}{1.2} 
\setlength{\tabcolsep}{4pt} 
\begin{tabular}{ll|cccccc}
\toprule
\textbf{Method} & \textbf{Avg.} & \textbf{FigureQA} & \textbf{DVQA-E} & \textbf{DVQA-H} & \textbf{PlotQA} & \textbf{ChartQA} & \textbf{ChartBench} \\
\midrule
Random & 21.3 & 50.0 & 25.8 & 25.6 & 8.9 & 12.8 & 4.8 \\
CLIP & 25.5 & 48.6 & 28.9 & 27.2 & 22.1 & 18.8 & 7.4 \\
\rowcolor[HTML]{F5F5F5} + Fine-tuning & 
41.5 {\tiny \textcolor[HTML]{228B22}{$\uparrow$ 16.0}} & 
64.4 & 54.9 & 53.9 & 42.4 & 23.7 & 9.5 \\
\rowcolor[HTML]{F5F5F5} + Neg. Cap. & 
\textbf{51.4} {\tiny \textcolor[HTML]{228B22}{$\uparrow$ 25.9}} & 
\textbf{82.0} & \textbf{65.2} & \textbf{61.0} & \textbf{54.1} & \textbf{29.7} & \textbf{16.2} \\
\bottomrule
\end{tabular}
\end{center}
\end{table*}
\subsection{CLIP Evaluation}
\label{sec:clip-eva}

For CLIP evaluation, we implement an Image-to-Text Retrieval task. Specifically, given an input image, the task is to retrieve the correct caption along with several hard negative ones.
The hard negative captions are specifically crafted to resemble the positive captions while being incorrect,
as described in the later~\textsection\ref{sec:hard-negative-construct}.
This retrieval evaluation is performed using the test sets from five chart-related datasets: FigureQA~\citep{kahou2017figureqa}, DVQA~\citep{kafle2018dvqa}, PlotQA~\citep{methani2020plotqa}, ChartQA~\citep{masry-etal-2022-chartqa}, and ChartBench~\citep{ChartBench}.

 We select the CLIP-ViT-L/14-336px~\citep{radford2021learning} model in our study, as its vision model is widely used in LVLMs such as InstructBLIP, LLaVA and LLaVA-Phi~\citep{dai2023instructblip,liu2024visual,liu2024improved,zhu2024llava}.  The retrieval evaluation results are presented in Table~\ref{tab:original-clip-eval}.
\paragraph{Original CLIP Exhibits Poor Retrieval Performance} While prior research has demonstrated that the original CLIP model achieves over 70\% accuracy on ImageNet classification, its retrieval performance on chart-related datasets is notably poor, with results approaching random guessing on benchmarks such as FigureQA and DVQA. This can be attributed to the fact that the original CLIP model, pretrained on web-crawled image-caption corpora, contains limited high-quality chart-related data. The poor retrieval accuracy is often interpreted as a sign of information loss in the encoded images~\citep{kamath2023whats,tong2024eyes}, suggesting the vision encoder bottleneck. However, as we will discuss later in \textsection\ref{sec:not-imply}, we further study it and find that low retrieval accuracy does not necessarily imply information loss.

\subsection{CLIP Improvement}
Observing the poor performance of the original CLIP, we explore methods to improve the chart understanding capabilities of CLIP. 
The first approach we try is to continue training CLIP on chart images with the original CLIP loss.
Inspired by NegCLIP~\citep{negclip2022}, which demonstrated that CLIP's failures may stem from learning shortcuts during training, we further implement another variant that incorporates hard negative samples into our training process. The hard negative captions help push the model to learn more discriminative features. Our strategies for constructing these hard negatives will be detailed in the following section \textsection\ref{sec:hard-negative-construct}. 



For training data, we exclude reasoning-type questions from the PlotQA dataset, as they are not suitable for CLIP training and deviate from our primary objective of analyzing CLIP's impact on LVLM's perceptual capabilities. In addition to the mentioned chart-related datasets, we incorporate additional datasets such as CLEVR~\citep{johnson2017clevr}, MapQA~\citep{chang2022mapqa}, and VQAv2~\citep{goyal2017making}, resulting in a training set of approximately 8 million samples. Detailed statistics of the training data are provided in Appendix ~\ref{app:statistics-data}.
Since most of these datasets consist of question-answer pairs, we utilize Llama3-8B-Instruct~\citep{dubey2024llama} to convert the question-answer pairs into assertive sentences, which are used as training and evaluation captions.

\subsection{Constructing Hard Negative Captions }
\label{sec:hard-negative-construct}
\citet{negclip2022} introduced NegCLIP by perturbing word order to construct hard negative captions, forcing CLIP to enhance relational understanding. Similar approaches have been applied to the fine-grained conceptual understanding of color, object, location, and size~\citep{rosch-etal-2024-enhancing}. In this work, we adapt the NegCLIP methodology to the domain of chart understanding.
The process begins by synthesizing incorrect answers, which are then converted into assertive captions using LLama3-8B-Instruct. These incorrect captions are used as hard negatives to compel CLIP to better understand and distinguish between relevant chart information.

During the synthesis of incorrect answers, we employ several strategies. 
For binary answers, we systematically flip responses (e.g., changing ``yes'' to ``no''). For numerical answers in datasets like PlotQA, we programmatically generate incorrect values by introducing error ranges between 5\% and 80\% of the ground truth, as Figure~\ref{ex:caption_hard_negative} shows. For questions about chart titles, like in PlotQA, LLama3-8B-Instruct generates plausible but incorrect responses.
Further details of the hard negative captions for all datasets are shown in Appendix~\ref{app:hard-negative}.





\begin{figure}[htbp]
    \centering
    \begin{tcolorbox}[
        enhanced,
        colback=darkgreen!10!white,
        colframe=darkgreen!70!black,
        arc=4mm,
        boxrule=1pt,
        fonttitle=\bfseries\color{white},
        title=Example: Caption and Hard Negative Caption.,
        width=0.48\textwidth
    ]
\textbf{Caption:}  
``The number of anaemic children in Malawi in 1991 was 76.3\%.''

\textbf{Hard Negative Caption:}  
``The number of anaemic children in Malawi in 1991 was 40.6\%.''
    
    \end{tcolorbox}
    \caption{An example of Caption and hard negative caption.}
    \label{ex:caption_hard_negative}
\end{figure}


\subsection{Performance of Enhanced CLIP} 
As in previous experiments, we use the CLIP-ViT-L/14-336px~\citep{radford2021learning} model. The model is trained with a batch size of 64, a learning rate of $5 \times 10^{-6}$, for 3 epochs on our collected training data, which consists of approximately 8 million samples. The retrieval evaluation results are also presented in Table~\ref{tab:original-clip-eval}.

\begin{figure} [t]
    \centering
    \includegraphics[width=0.4\textwidth]
    {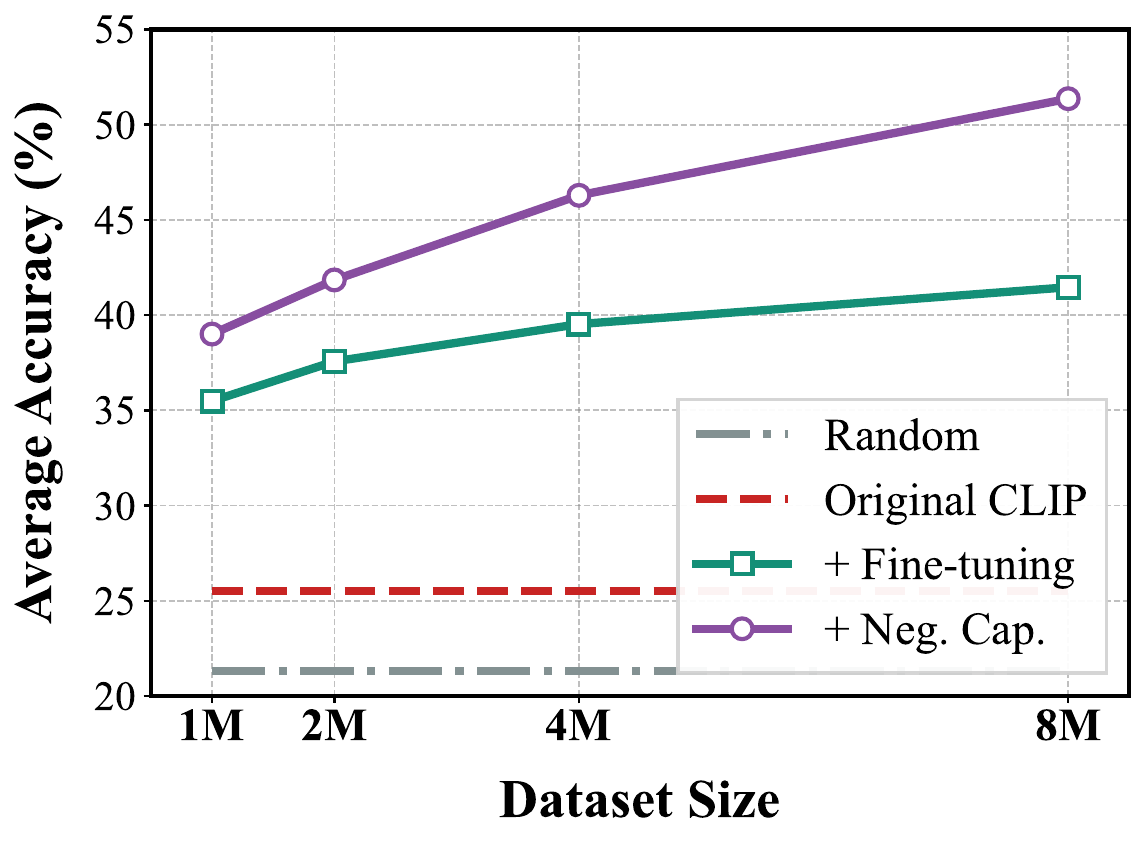}
 
    \caption{
    CLIP retrieval accuracy while scaling training data size, on the average of five datasets: FigureQA, DVQA-Easy\&Hard, PlotQA, ChartQA, Chartbench.
    }
    \vspace{-15pt}
    \label{fig-clip-scal}
\end{figure}

\paragraph{Fine-tuned CLIP Significantly Improves Retrieval Accuracy} Compared to the original CLIP, both fine-tuned models (with and without hard negatives) show significant improvements in retrieval performance. Furthermore, NegCLIP (CLIP fine-tuned with neg. cap.) achieves the largest improvement, surpassing 26 points across these datasets. 
Training data scaling experiments, shown in Figure~\ref{fig-clip-scal}, illustrate that the performance of fine-tuned CLIP improves steadily with larger training datasets, while NegCLIP consistently outperforms the other models.
We conclude that incorporating hard negative captions effectively forces CLIP to learn more accurate and relevant chart information, similar to the success of NegCLIP in previous works~\citep{negclip2022,rosch-etal-2024-enhancing}. 

While prior research has identified limitations of CLIP in handling subtle visual patterns~\citep{tong2024eyes} and spatial reasoning~\citep{kamath2023whats}, often attributing these issues to its inductive biases, our improvements in classifying subtle chart type features demonstrate that such limitations can be mitigated through data-centric contrastive learning.

\section{The Extraction Bottleneck: Connecting CLIP  to LVLM}

\begin{table*}[t]
\caption{Evaluation accuracy of LLaVA-v1.5-13B, LLaVA-Chart-13B and LLaVA-Chart-Phi based on different CLIPs. The first result row, labeled "LLaVA," corresponds to LLaVA-v1.5-13B without chart-specific tuning. ``Binary'' indicates tasks with Yes/No answers. ``Frozen'' and "Unfrozen" refer to whether the CLIP model is frozen during LLaVA training. ``FT.CLIP'' represents the fine-tuned CLIP without hard negative captions, while ``NegCLIP'' refers to the CLIP trained with hard negative captions. 
The \colorbox{LightCyan}{$\Delta$} rows report per-benchmark performance gains of Unfrozen-NegCLIP compared to Unfrozen-CLIP.}
\label{main-res}
\vskip 0.15in
\begin{center}
\begin{small}
\tabcolsep=2.7pt
\begin{tabular}{llc|cccccccccccc}
\toprule
\multirow{2}{*}{\textbf{VLM}} & \multirow{2}{*}{\makecell[c]{\textbf{Vision} \\ \textbf{Encoder}}} & \multirow{2}{*}{\textbf{Avg.}} & \textbf{FigureQA} &  \multicolumn{2}{c}{\textbf{DVQA}} & \textbf{PlotQA} & \textbf{ChartQA} &  \multicolumn{2}{c}{\textbf{ChartBench}} & \multicolumn{2}{c}{\textbf{MathVista}}  & {\textbf{ChartX}} \\
\cmidrule(lr){4-4} \cmidrule(lr){5-6} \cmidrule(lr){7-7}  \cmidrule(lr){8-8} \cmidrule(lr){9-10} \cmidrule(lr){11-12} \cmidrule{13-13}
& &  & \textbf{Binary} & \textbf{Easy} & \textbf{Hard} & \textbf{QA} & \textbf{QA} & \textbf{Binary} & \textbf{QA} & \textbf{FQA} & \textbf{ALL} & \textbf{QA} \\
\midrule
\multirow{3}{*}[45pt]{\rotatebox{90}{\parbox{2.5cm}{\textbf{CLIP}}}} & CLIP & - & 48.6 & 28.9 & 27.2 & 22.1 & 18.8 & - & 7.4 & - & - & -\\
\textbf{} & FT.CLIP & - & 64.4 & 54.9 & 53.9 & 42.4 & 23.7  & - & 9.5 & - & - & -\\
\textbf{} & NegCLIP & - & 82.0 & 65.2 & 61.0 & 54.1  & 29.7 & - & 16.2 & - & - & - \\
\midrule
\textbf{LLaVA} & Frozen-CLIP & 25.9 & 51.2 & 25.8 & 25.3 & 12.6 & 18.3 & 53.0 & 9.7 & 23.1 & 27.0 & 12.7\\
\cdashline{1-13}
\multirow{6}{*}[20pt]{\rotatebox{90}{\parbox{2.5cm}{\textbf{LLaVA-\\Chart-13B}}}} &  Frozen-CLIP & 53.2 & 78.4 & 79.9 & 75.4 & 41.7 & 53.0 & \textbf{73.4} & 26.4 & 49.4 & 34.0 &20.1\\
& Unfrozen-CLIP & 53.6 & 78.9 & 79.7 & 74.9 & 41.7 & 53.1 & 73.2 & \textbf{27.8} & 50.9  & \textbf{36.1} &19.6\\ \cdashline{2-13}
&Frozen-FT.CLIP & 54.8 & 83.8 & 84.3 & 78.7 & 43.8 & 54.3  & 73.1 & 26.3 &  48.0 & 34.4  & 21.2 \\
& Unfrozen-FT.CLIP & 55.2  & 83.4 & 84.4 & 78.9 & 44.1 & 54.6 & 73.2 & 26.9 & 49.4  & 35.7 &20.8  \\
&  Frozen-NegCLIP & 56.0 & \textbf{86.2}  & 86.1 & \textbf{80.9}  & 44.8 & 54.9 & 72.1  & 27.1  & 52.0 & 34.6  & \textbf{21.5} \\
& Unfrozen-NegCLIP & \textbf{56.2}  & 86.0 & \textbf{86.3} & 80.7 & \textbf{45.1} & \textbf{55.0} & 72.8 & 26.9 & \textbf{52.4} & 35.4 & 21.4  \\
\rowcolor{LightCyan}
& $\Delta$ over Unfrozen-CLIP & +2.6 & +7.1 & +6.6 & +5.8 & +3.4 & +1.9 & -0.4 & -0.9 & +1.5 & -0.7 & +1.8\\
\midrule
 \multirow{6}{*}[20pt]{\rotatebox{90}{\parbox{2.5cm}{\textbf{LLaVA-\\Chart-Phi}}}}  & Frozen-CLIP & 49.4
 & 72.1 & 76.1 & 70.6 &38.9 & 48.0 & 70.9 & 23.3 & 43.5 & 33.4 &17.5\\
& Unfrozen-CLIP &  49.3 & 71.3 & 76.7 & 70.5 & 38.5 & 48.1 & 71.7 & 23.8 & 40.5 & 33.7 &18.1\\ \cdashline{2-13}
 & Frozen-FT.CLIP & 52.0 & 79.3 & 81.8  & 75.2  & 41.7 & 49.7 &  \textbf{71.8} & 23.3  & 45.4  & 34.2  &17.8 \\
&  Unfrozen-FT.CLIP & 51.7 & 78.6  & 81.7 & 74.8  & 41.5 & 49.4 & 71.1 &  23.5 & 46.1  & 33.1  &17.5 \\
&  Frozen-NegCLIP & 54.1 & 85.0 & \textbf{85.0} & \textbf{78.3} & 42.5  & \textbf{51.3} &  71.2& 24.2 & 49.4  & 34.9 & 19.0\\
&  Unfrozen-NegCLIP & \textbf{54.3}  & \textbf{85.1} & 84.9 & 77.6 & \textbf{42.6} & 51.0 & 70.9 & \textbf{24.8} & \textbf{50.6} & \textbf{35.6}  & \textbf{19.5}  \\
\rowcolor{LightCyan}
& $\Delta$ over Unfrozen-CLIP & +5.0 & +13.8 & +8.2 & +7.1 & +4.1 & +2.9 & -0.8 & +1.0 & +10.1 & +1.9 & +1.4\\
\bottomrule
\end{tabular}
\end{small}
\end{center}
\vskip -0.1in
\end{table*}

Upon finishing our study of CLIP for the vision encoder bottleneck, we shift our focus to the extraction bottleneck to understand how these CLIP models impact LLaVAs. Having observed the poor performance of the original CLIP and the improved performance of fine-tuned CLIPs, we aim to answer  two questions: 

\begin{itemize}[leftmargin=*]
    \item \textit{Does the failure of CLIP retrieval cause the failure of LLaVAs that are based on it?}
    \item \textit{What is the impact of enhanced CLIPs on the performance of LLaVAs?}
\end{itemize}

\subsection{Experimental Setup}
\label{sec:training-setup}
\paragraph{Training Setup} Following LLaVA-v1.5-13b~\citep{liu2024improved} and LLaVA-Phi~\citep{hanoona2024LLaVA++}, we use Vicuna-13b~\citep{vicuna2023} or the Phi-3-mini~\citep{abdin2024phi} of 3.8B parameters as the base LLM and employ a two-layer MLP connector to map CLIP's image embeddings into the LLM's input space. Our training process consists of three stages. First, we pretrain the connector on 558K image-caption pairs from the LLaVA training dataset, keeping both the CLIP vision encoder and the LLM fixed. In the second stage, we conduct visual instruction tuning on 665K instruction samples, also derived from the LLaVA dataset. Finally, in the third stage, we perform chart-specific tuning on a dataset of 250K chart samples, including FigureQA, DVQA, PlotQA, ChartQA, and Chart2Text~\citep{kantharaj-etal-2022-chart}, resulting in the LLaVA-Chart-13B and LLaVA-Chart-Phi models. In both the second and third stages, we explore two strategies: freezing or unfreezing the CLIP vision encoder.

\paragraph{Evaluation Setup}
We sample 25K examples separately  from the test sets of FigureQA, DVQA, and PlotQA for evaluation. For FigureQA and DVQA, we use exact match accuracy as the evaluation metric. For numerical answers in PlotQA, we adopt a relaxed correctness criterion, considering a prediction correct if it falls within 5\% of the ground truth, following prior works~\citep{methani2020plotqa}.
For ChartQA, we use its 2.5K test set and apply the same relaxed correctness criterion for numerical answers. Similarly, for ChartBench, we focus on QA tasks and split the dataset into two subtasks: binary QA (Yes/No answers) and 2.1K numerical QA samples, applying relaxed correctness for the latter.
Additionally, to evaluate generalization performance, we include the MathVista benchmark~\citep{lu2024mathvista} and ChartX~\citep{xia2024chartx}.

\subsection{Poor Retrieval Performance Does Not Imply Limited Information Encoding}
\label{sec:not-imply}
Our experimental results (Table~\ref{main-res}) show that LLaVA, without the third-stage chart-specific tuning, performs poorly on chart benchmarks, achieving lower accuracy than the original CLIP retrieval performance. After chart-specific tuning, LLaVA based on the original CLIP can learn these chart tasks successfully, even when the CLIP is frozen, indicating the improved extraction ability. For instance, LLaVA-Chart-13B achieves 78\% accuracy on FigureQA, despite its CLIP nearly random retrieval accuracy on the same dataset in Table~\ref{tab:original-clip-eval}. Moreover, we observe that unfreezing the vision encoder provides only a minor improvement.
Importantly, the success of the original CLIP-LLaVA training suggests that the original CLIP is not ``blind''; poor retrieval performance does not necessarily indicate a lack of encoded information within CLIP's image embeddings. Prior works~\citep{tong2024eyes,kamath2023whats} have likely overemphasized the concept of CLIP's blindness.
We hypothesize that retrieval accuracy primarily reflects the linear properties of CLIP’s image and text embeddings, as similarity computation in retrieval tasks relies on cosine similarity or dot product, which are inherently linear operations. However, when integrated into an LVLM, the LLM component—acting as a more powerful information processor—can extract and utilize non-linear features from CLIP’s image embeddings.
Similar observations have been reported in recent work~\citep{li2024erroneous}, indicating that retrieval accuracy may be an inadequate proxy for assessing the vision encoder bottleneck.

To further validate this conclusion, we conducted an ablation study by training LLaVA with randomly initialized CLIP weights. Notably, the model failed to converge during the final chart-specific fine-tuning stage (details in the Appendix~\ref{app:rand-clip}). This confirms that the original CLIP, despite their as poor as random retrieval performance, provides crucial visual information for successful LVLM training.

\begin{figure} [t]
    \centering
    \includegraphics[width=0.4\textwidth]
    {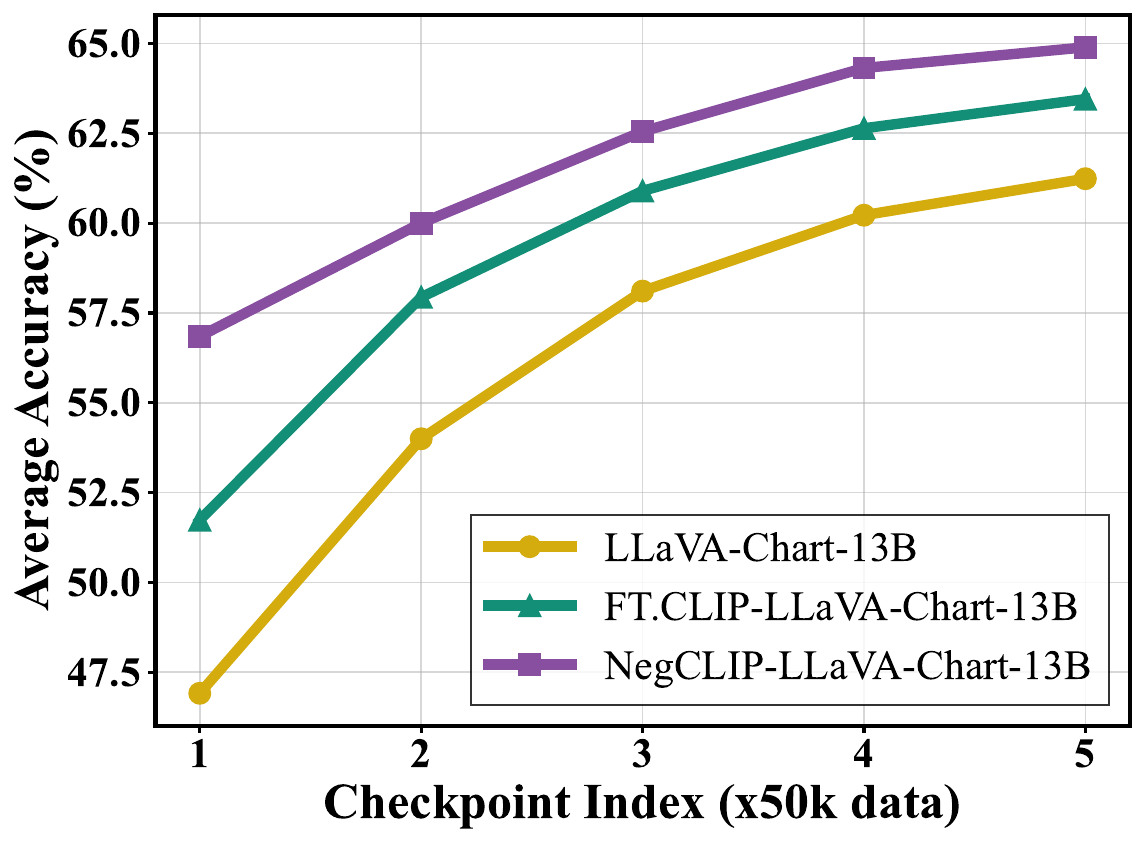}
 
    \caption{LLaVA training data scaling results, averaged over five datasets: FigureQA, DVQA-E\&H, PlotQA, ChartQA, and ChartBench, for LLaVAs based on different CLIP vision encoders (the original CLIP, FT.CLIP, and NegCLIP).}
    \vspace{-15pt}
    \label{fig-chart—mix-scal}
\end{figure}

\subsection{Enhanced CLIPs Elevate LVLMs Performance}

The success of the original CLIP-LLaVA makes the relationship between CLIP and LLaVA performance less intuitive. To explore the impact of enhanced CLIPs on LVLMs, we conducted the same training experiments using these enhanced CLIPs as vision encoders. Our findings reveal that LLaVAs based on enhanced CLIPs consistently achieve significantly better performance.
Consistent with the results from the CLIP evaluation (\textsection\ref{sec:clip-eva}), NegCLIP-LLaVAs demonstrate the best performance across most benchmarks. Specifically, for in-distribution datasets, NegCLIP-LLaVAs achieve improvements exceeding 5 absolute points on FigureQA, DVQA, and PlotQA. Additionally, the improvements observed on MathVista and ChartX highlight the generalization capability of LLaVAs built upon our enhanced CLIP models. On average, compared to the Unfrozen-CLIP baseline, models based on NegCLIP exhibit notable gains: LLaVA-Chart-13B improves by 2.6 absolute points, while LLaVA-Chart-Phi achieves an even larger improvement of 5.0 absolute points.
Additionally, data scaling experiments during the third-stage chart-specific tuning, illustrated in Figure~\ref{fig-chart—mix-scal}, demonstrate consistent performance improvements with increased training data. Across the scaling process, NegCLIP-LLaVAs consistently achieve the highest performance.

These results confirm that while chart-specific tuning helps mitigate the extraction bottleneck, addressing the vision encoder bottleneck remains critical for achieving greater performance gains. We hypothesize that enhanced CLIP encodes more salient information in its image representations, thereby making LVLM training easier. Further insights are discussed in \textsection\ref{sec:rethinking}.

\section{Scaling Chart Understanding Tuning}
\label{sec:large-scal}

In this section, we scale up task-specific training data to 800K samples per dataset to fully mitigate the extraction bottleneck, enabling the performance of LVLMs to directly reflect the extent of information encoded by CLIP.


Specifically, we conduct training for both CLIP and LLaVA using the DVQA and PlotQA datasets separately, leading to the two specialized models: LLaVA-PlotQA and LLaVA-DVQA. For CLIP training, we utilize a total of 2 million samples from DVQA and 3 million samples from PlotQA. We still incorporate both standard training data and hard negative variants, following the hard negative generation strategy and hyperparameter configuration detailed in Section~\ref{sec:hard-negative-construct}.
For LLaVA training, we adhere to the three-stage training process using the LLaVA-v1.5-13B model, as outlined in Section~\ref{sec:training-setup}. In the third and final chart-specific tuning stage, we train LLaVA models using 800K samples from each dataset separately, allowing us to systematically investigate the performance ceiling under this setting. 



\begin{figure*} [t]
	\centering
     \subfloat[\label{fig-plotqa-scal-ana}PlotQA]{
    \includegraphics[scale=0.25]{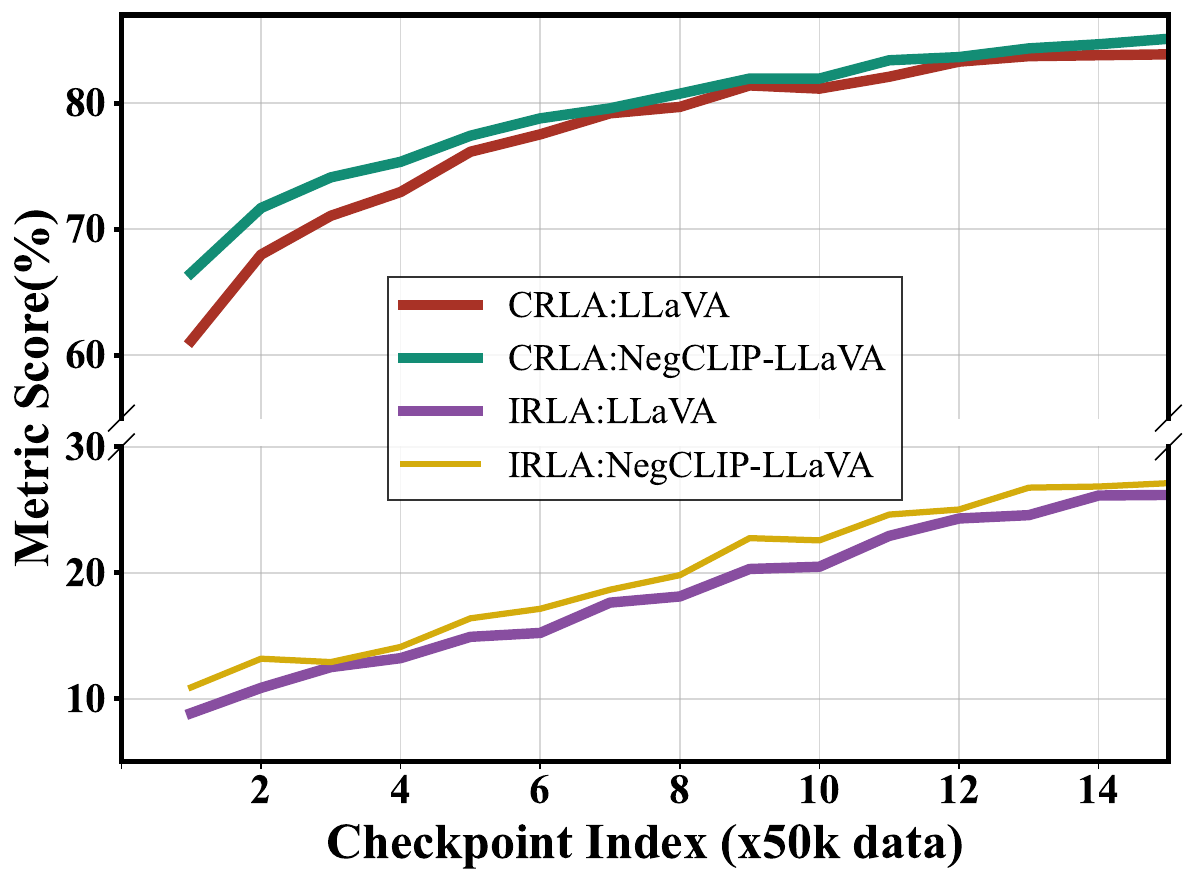}}
	\subfloat[\label{fig-dvqa-easy-scal-ana}DVQA-Easy]{
		\includegraphics[scale=0.25]{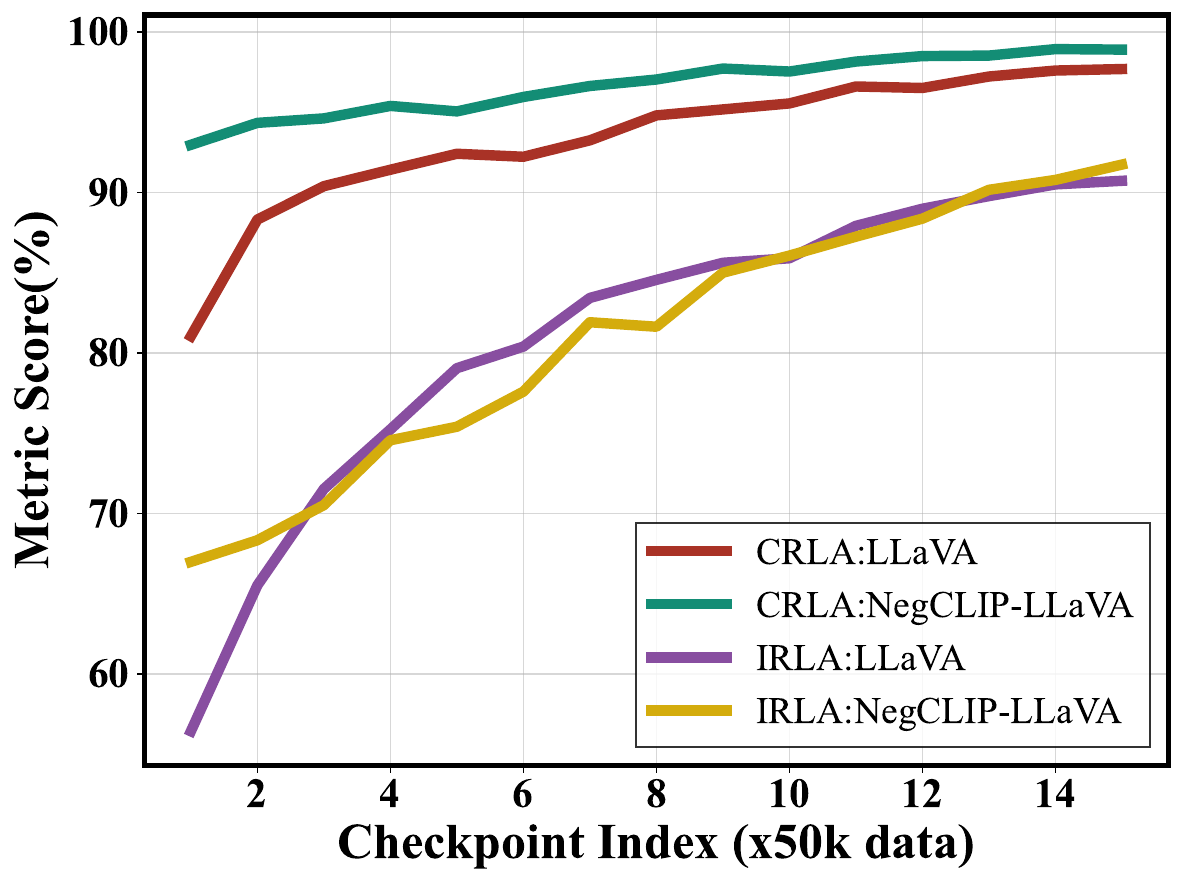}}
	\subfloat[\label{fig-dvqa-hard-scal-ana}DVQA-Hard]{
		\includegraphics[scale=0.25]{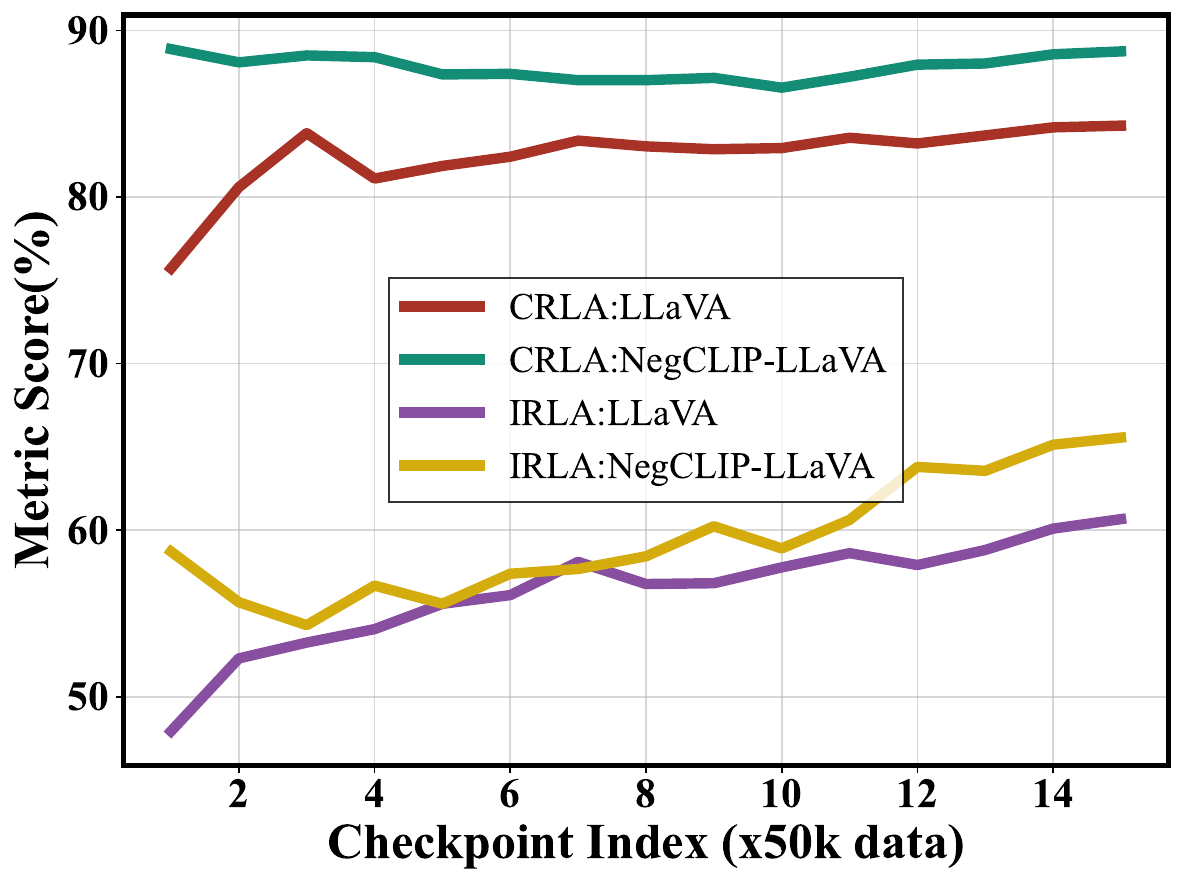}}
  
  \caption{Analysis of large-scale LLaVA SFT data scaling on PlotQA and DVQA-Easy\&Hard, evaluating two metrics: Correct-Retrieval LLaVA Accuracy (CRLA) and Incorrect-Retrieval LLaVA Accuracy (IRLA) for both the original CLIP-LLaVA and NegCLIP-LLaVA.}
	\label{fig-scal-single-ana} 
    \vspace{-15pt}
\end{figure*}

\begin{table}[t]
\caption{Performance results on DVQA-Easy, DVQA-Hard, and PlotQA for different CLIP vision encoder-based LLaVA-Specific models trained on large-scale chart-specific datasets (800K samples from either DVQA or PlotQA). Improvements (\textcolor[HTML]{228B22}{$\uparrow$}) are shown relative to the LLaVA-Specific baseline. }
\label{tab:scal-chart}
\begin{center}
\begin{small}
\renewcommand{\arraystretch}{1.2} 
\setlength{\tabcolsep}{2pt} 
\begin{tabular}{l|lllccc}
\toprule
\textbf{Model} & \textbf{DVQA-E} & \textbf{DVQA-H} & \textbf{PlotQA}  \\
\midrule
LLaVA-Specific & 95.1 & 74.4 & 58.9 \\
\rowcolor[HTML]{F5F5F5} FT.CLIP-LLaVA-Specific & 95.3 & 76.6 & 59.5\\
\rowcolor[HTML]{F5F5F5} NegCLIP-LLaVA-Specific & \textbf{96.0} {\tiny \textcolor[HTML]{228B22}{$\uparrow$ 0.9}} & \textbf{78.2} {\tiny \textcolor[HTML]{228B22}{$\uparrow$ 3.8}} & \textbf{60.0} {\tiny \textcolor[HTML]{228B22}{$\uparrow$ 1.1}}\\
\bottomrule
\end{tabular}
\end{small}
\end{center}
\vspace{-5pt}
\end{table}

\subsection{Experimental Results}

As shown in Table~\ref{tab:scal-chart}, scaling the training data to 800K samples per dataset significantly improves performance on specific tasks by further mitigating the extraction bottleneck.
Despite being trained on much larger task-specific datasets, our enhanced CLIPs still achieve a higher LVLM performance ceiling. Notably, NegCLIP-LLaVA surpasses its original CLIP-based counterparts by an additional 1 absolute point on PlotQA and DVQA-Easy, and 4 absolute points on DVQA-Hard. Detailed performance scores throughout the training process are provided in Appendix~\ref{app:large-scale}.
The superior performance observed after large-scale instruction tuning suggests that enhanced CLIPs encode more useful information, thereby contributing to a higher LVLM performance ceiling.




\subsection{Statistics on CLIP and LLaVA Behavior}
To discover deeper insights into how CLIP retrieval capabilities translate into LLaVA task-specific performance, we analyze the statistics between CLIP retrieval correctness and LLaVA task correctness across these scaling experiments. Specifically, we use NegCLIP and examine two metrics: (1) \textbf{Correct-Retrieval LLaVA Accuracy (CRLA)}: LLaVA accuracy when NegCLIP retrieves samples correctly. (2) \textbf{Incorrect-Retrieval LLaVA Accuracy (IRLA)}: LLaVA accuracy when NegCLIP retrieves samples incorrectly. We analyze these two metrics using the original CLIP-LLaVA and the NegCLIP-LLaVA which are fine-tuned on the large-scale PlotQA or DVQA dataset. The results are illustrated in Figure~\ref{fig-scal-single-ana}.

\paragraph{Results}  
The analysis reveals that CRLA is significantly higher than IRLA, indicating that samples correctly retrieved by NegCLIP are easier for LLaVA to learn. Moreover, during the early stages of instruction training, NegCLIP-LLaVA exhibits a markedly higher CRLA than the original CLIP-LLaVA, which is the primary source of the performance gap.  This intuitive ``CLIP Can, LLaVA Can'' observation suggests that NegCLIP encodes more salient features, enabling LLaVA to learn faster and more effectively.

As training data scales, the difference in CRLA between NegCLIP-LLaVA and original CLIP-LLaVA decreases, reflecting a narrowing performance gap. Meanwhile, for both the original CLIP-LLaVA and NegCLIP-LLaVA, IRLA steadily improves, suggesting that LLaVA can progressively leverage additional non-linear information beyond what is explicitly indicated by retrieval accuracy.



\subsection{Rethinking Information Encoding in CLIP}
\label{sec:rethinking}
Finally, we reconsider how CLIP encodes information in relation to its retrieval accuracy. Retrieval accuracy primarily reflects the linear properties of the image embeddings due to the similarity in retrieval task operates within a linear space.
However, when the CLIP vision encoder is integrated into LLaVAs, the LLM component, being a more powerful and flexible information extractor, can extract and utilize non-linear features embedded in CLIP's image representations. This means that certain aspects of the encoded information, which might not directly contribute to retrieval accuracy, can still be used for downstream tasks.

Therefore, poor retrieval accuracy does not necessarily imply a loss of crucial encoded information. Instead, through mitigating the vision encoder bottleneck, the enhanced CLIP makes its encoded information more salient, i.e. linear,  as evidenced by the improved retrieval accuracy.
 At the same time, the more salient image embeddings make it easier for LLaVA to learn, thereby enabling the LLaVA to converge faster and achieve higher performance in downstream tasks.

\section{Conclusion}
This study explores the perception bottlenecks of LVLMs for chart understanding through the vision encoder bottleneck and the extraction bottleneck. We address the vision encoder bottleneck through chart-tailored contrastive learning. Furthermore, LVLMs built on these improved CLIP models demonstrate substantial performance gains. Our findings emphasize how the capabilities of CLIP influence LLaVA's downstream task performance, offering valuable insights into understanding CLIP information encoding.


\section*{Limitations}
Our work aims to deepen the understanding of the vision encoder effect on LVLMs for chart understanding. However, there are some limitations.

First, our goal is not to develop a state-of-the-art LVLM for chart understanding, as many advanced models are either closed-source or prohibitively expensive to reproduce. Instead, our work aims to provide a deeper understanding of LVLMs by analyzing the vision encoder bottleneck and the extraction bottleneck of the language model.
Second, due to computational constraints, our experiments are limited to a single vision encoder: CLIP-ViT-L/14-336px. Investigating other vision encoder variants, such as SigLIP~\citep{zhai2023sigmoid}, remains for future research.

While our study primarily focuses on chart understanding, the success of NegCLIP training and NegCLIP-LLaVA suggests broader applicability beyond this domain, which we leave for future exploration.

\section*{Acknowledgements}
This project is partially supported by NSFC Grant 62306177.

\bibliography{custom}

\appendix
\section{Details of CLIP Training Data }
\subsection{Statistics of Training Data}
In Table~\ref{app:statistics}, we present the statistics of the datasets included in the CLIP training. Here, we upsampling ChartQA and ChartBench to maintain data balance. To ensure balanced data distribution, we upsampled the ChartQA and ChartBench datasets.
\label{app:statistics-data}

\begin{table*}[h] 
\centering
\caption{The statistics of datasets used for CLIP training. \# Images is the total number of images for each dataset. \# Captions is the total number of captions for each dataset in the final mixture.}
\label{app:statistics} 
\begin{tabular}{lcc} 
\toprule
\textbf{Dataset} & \textbf{\# Images} & \textbf{\# Captions} \\
\midrule
FigureQA~\citep{kahou2017figureqa} & 99,992 & 1,000,000\\ 
DVQA~\citep{kafle2018dvqa} & 200,000 &  2,000,000\\
PlotQA~\citep{methani2020plotqa} & 157,044 &  2,000,000 \\
ChartBench~\citep{ChartBench} & 133,248   & 568,475 \\
Chart2text~\citep{kantharaj-etal-2022-chart} & 26,961 & 87,946 \\
ChartQA~\citep{masry-etal-2022-chartqa} & 18,317 & 169,030 \\
WikiSQL~\citep{zhong2017seq2sql} & 74,989 & 288,893 \\
CLEVR~\citep{johnson2017clevr} & 70,000 &  699,989 \\ 
DocVQA~\citep{mathew2021docvqa} & 10,189 &  39,463 \\
OCR-VQA~\citep{mishra2019ocr} & 165,746 & 801,579 \\
MapQA~\citep{chang2022mapqa} & 12,470 & 151,536 \\
TextVQA~\citep{singh2019towards} & 21,953 &  34,601 \\
A-OKVQA~\citep{okvqa} & 16,539 & 17,056 \\
VQAv2~\citep{goyal2017making} & 82,772 & 443,756\\
\bottomrule
\end{tabular}
\end{table*}

\subsection{Details of Hard Negative Captions Construction}
\label{app:hard-negative}
To generate hard negative captions, we first apply specific strategies to produce incorrect answers and then use Llama3-8B-instruct to convert the question-answer pairs into assertive sentences as hard negative samples.

\paragraph{FigureQA:} Since FigureQA answers are binary (``Yes'' or ``No''), we construct hard negatives by flipping the correct answers.

\paragraph{DVQA:} For DVQA, we flip the binary answers (e.g., ``Yes'' to ``No'' and vice versa). For categorical answers (e.g., labels), we either randomly select another label from the chart or utilize Llama3-8B-instruct to generate a similar but incorrect label. For numerical and other types of answers, we consistently leverage Llama3-8B-instruct to produce plausible but incorrect alternatives.

\paragraph{PlotQA:} For numerical answers, we systematically generate incorrect values by introducing errors ranging from 5\% to 80\% of the ground truth. For non-numerical answers, we again rely on Llama3-8B-instruct to produce reasonable yet incorrect alternatives.

\paragraph{ChartBench:} The same strategies as used for PlotQA are applied to generate hard negative answers.

\paragraph{Chart2text:} We split the text descriptions into individual captions corresponding to the image. Then, we use Llama3-8B-instruct to modify the meaning of these captions, such as altering numerical values, to create hard negatives.

\paragraph{ChartQA:} The approach for ChartQA mirrors that of PlotQA, using similar strategies to generate hard negative answers.

\paragraph{Others:} For other datasets, we exclusively use Llama3-8B-instruct to generate incorrect answers.

\section{Details Experimental Results}
\subsection{Investigation into LLaVA-Random-CLIP}
\label{app:rand-clip}

In \textsection\ref{sec:not-imply}, we observed that LLaVA based on the original CLIP successfully learned chart-related tasks, even though the original CLIP exhibited poor, almost random retrieval accuracy. This raises an important question: is the visual information encoded by CLIP truly random? To address this, we conducted an ablation experiment by randomly initializing the CLIP weights and training a random-CLIP-based LLaVA to determine whether LLaVA can still successfully learn chart tasks in this scenario.

\paragraph{Experimental Setup:}
In this experiment, we used a randomly initialized CLIP while retaining the same three-stage training procedure for LLaVA as described in the paper. Specifically, we employed 800K FigureQA samples as the training data for the third stage.

\begin{figure*} [h]
    \centering
    \includegraphics[width=0.9\textwidth]
    {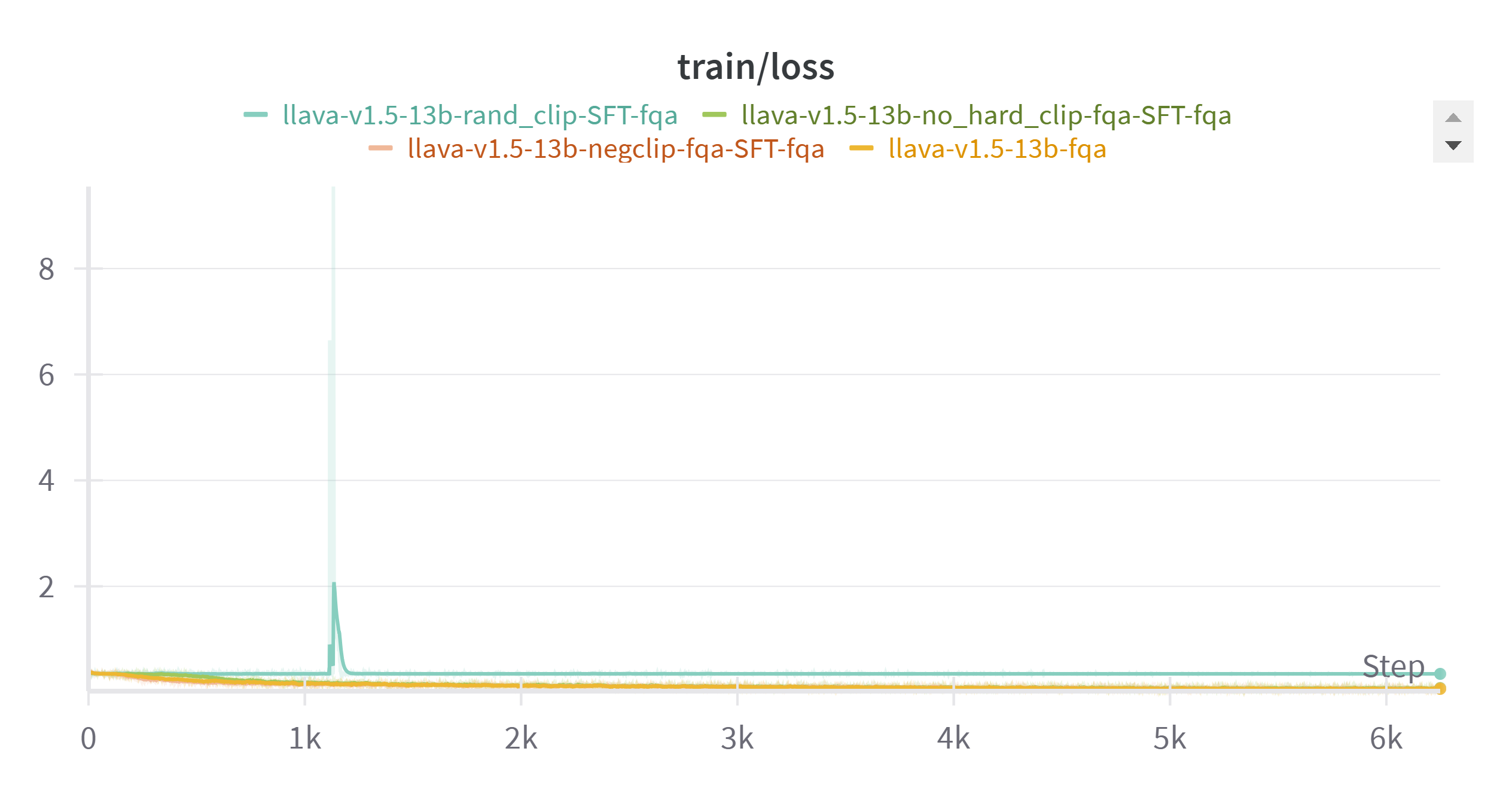}
 
    \caption{FigureQA instruction tuning loss of LLaVA-v1.5-13b based on different vision encoders.}
    \label{fig-rand}
\end{figure*}

\paragraph{Experimental Results:}
The results reveal that the training loss failed to converge during the final stage, as shown in the detailed loss plot (Figure~\ref{fig-rand}). These ablation results demonstrate that purely random information leads to the failure of LVLM learning. Moreover, the poor performance of the original CLIP does not imply that its encoded information is entirely random. In fact, the original CLIP still captures critical visual information, which is essential for the successful learning of LVLMs.

\subsection{Results of 800K Scaling Experiments}
\label{app:large-scale}
\begin{figure*} [h]
	\centering
     \subfloat[\label{fig-plotqa-scal}PlotQA]{
    \includegraphics[scale=0.25]{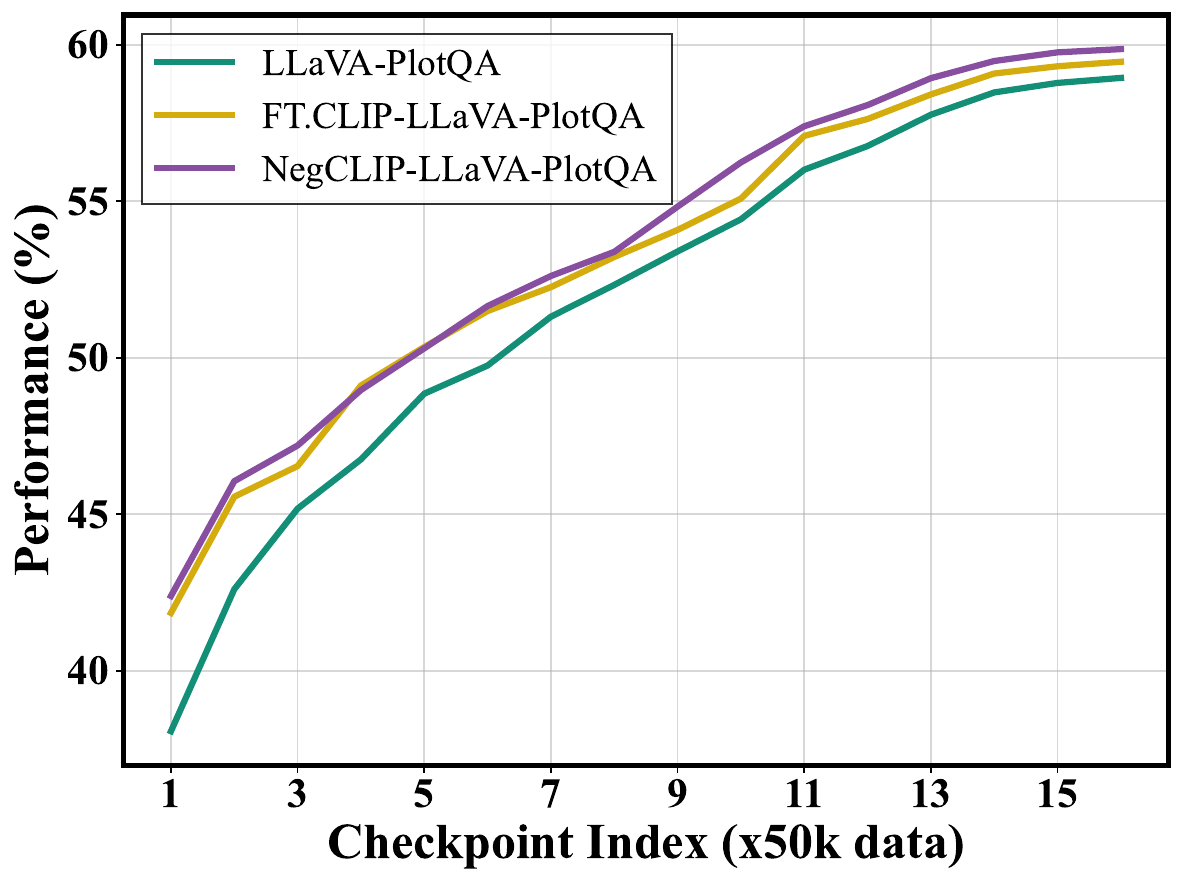}}
	\subfloat[\label{fig-dvqa-easy-scal}DVQA-Easy]{
		\includegraphics[scale=0.25]{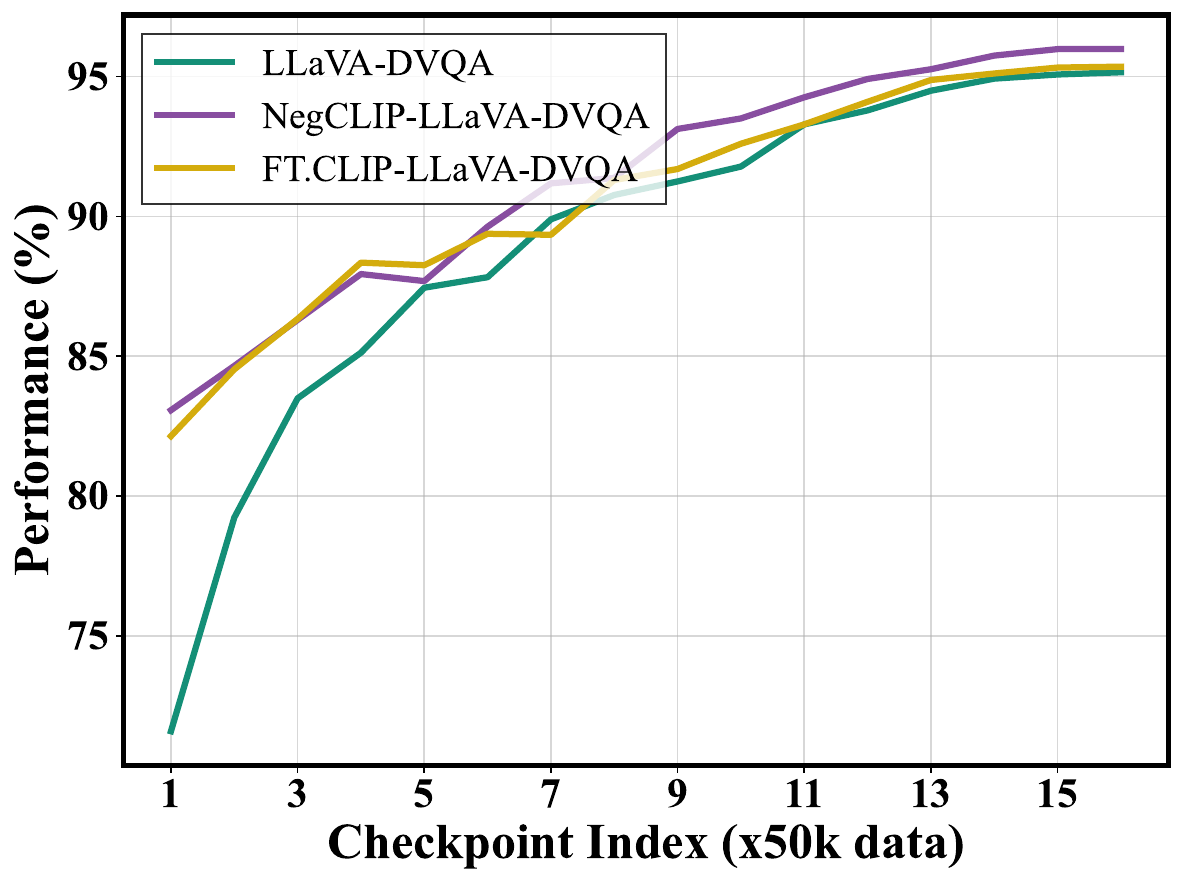}}
	\subfloat[\label{fig-dvqa-hard-scal}DVQA-Hard]{
		\includegraphics[scale=0.25]{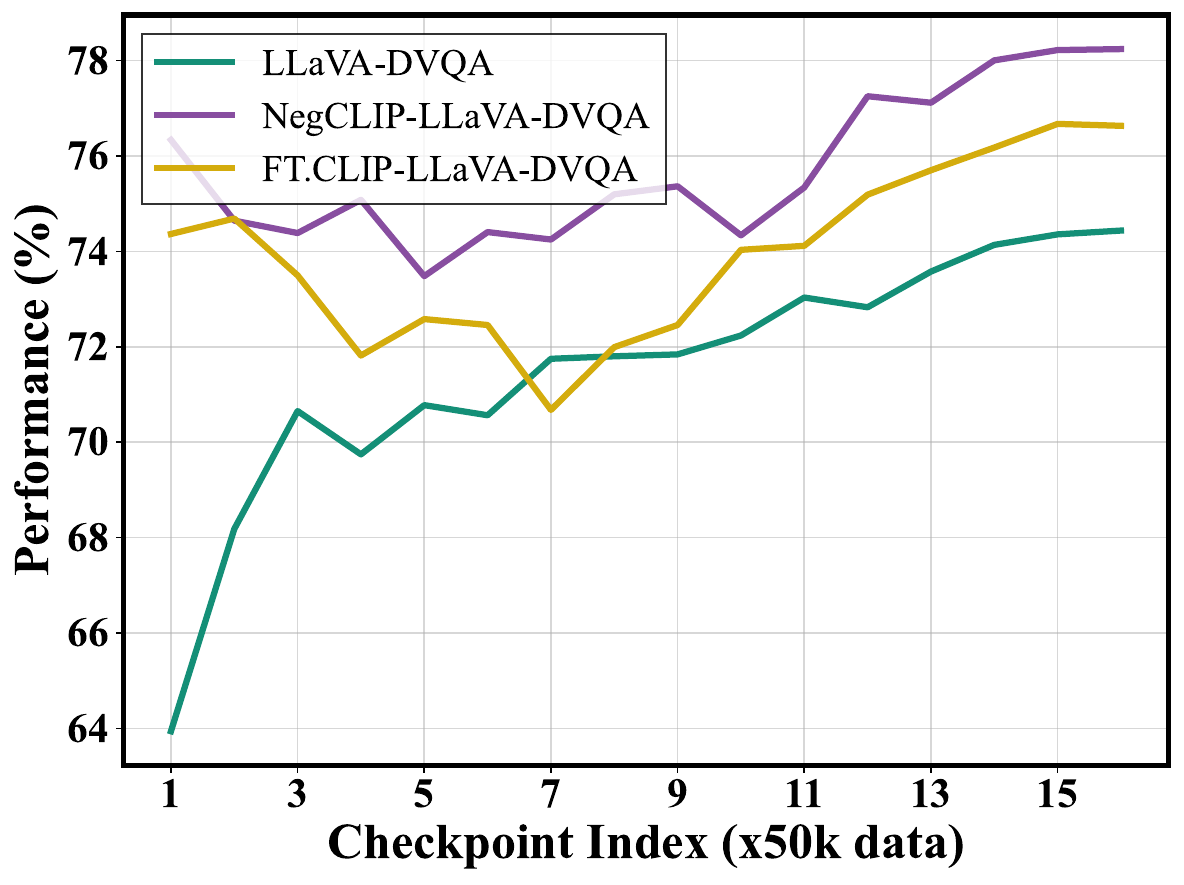}}
  \caption{The large LLaVA SFT data scaling results on PlotQA and DVQA-Easy\&Hard, for LLaVAs based on different CLIP vision encoders (the original CLIP, FT.CLIP, and NegCLIP).}
	\label{fig-scal-single} 
\end{figure*}

In \textsection~\ref{sec:large-scal}, we perform large-scale instruction tuning on 800K samples from the DVQA and PlotQA datasets separately. The evaluation performance throughout the training process is shown in Figure~\ref{fig-scal-single}. We observe that scaling up the training data results in steady improvements. Additionally, our enhanced CLIP-based LLaVA consistently achieves higher performance, indicating that the enhanced CLIP encodes more useful and salient information.



\end{document}